\documentclass[letterpaper]{article} 
\usepackage{aaai25}  
\usepackage{times}  
\usepackage{helvet}  
\usepackage{courier}  
\usepackage[hyphens]{url}  
\usepackage{graphicx} 
\usepackage{natbib}  
\usepackage{caption} 
\usepackage{algorithm}
\usepackage{algorithmic}

\usepackage{newfloat}
\usepackage{listings}

\usepackage{times}
\usepackage{soul}
\usepackage{url}
\usepackage[utf8]{inputenc}
\usepackage{caption}     
\usepackage{amsmath}
\usepackage{amsthm}
\usepackage{booktabs}
\usepackage{algorithm}
\usepackage{algorithmic}
\usepackage{subcaption}
\usepackage{natbib}
\usepackage{amsfonts}
\usepackage{array}
\usepackage{tikz}
\usepackage{booktabs}
\usepackage{tabularx}
\usepackage{makecell}
\usetikzlibrary{positioning, arrows.meta, shapes.geometric}
\usepackage{booktabs}
\usepackage{tabularx}
\usepackage{makecell}
\definecolor{lightyellow}{HTML}{FFF9C4}
\usepackage{textcomp}
\usepackage{url}
\usepackage{verbatim}
\usepackage{graphicx}
\usepackage{cite}
\usepackage{multirow}
\usepackage{upgreek, bm}

\DeclareCaptionStyle{ruled}{labelfont=normalfont,labelsep=colon,strut=off} 
\floatstyle{ruled}
\newfloat{listing}{tb}{lst}{}
\floatname{listing}{Listing}
\title{Dynamic Modes as Time Representation for Spatiotemporal Forecasting}
\author{
    Menglin Kong\textsuperscript{\rm 1},
    Vincent Zhihao Zheng\textsuperscript{\rm 1},
    Xudong Wang\textsuperscript{\rm 1}
Lijun Sun\textsuperscript{\rm 1}\thanks{Corresponding author}
}
\affiliations{
    \textsuperscript{\rm 1}Department of Civil Engineering, McGill University\\

\{menglin.kong, zhihao.zheng, xudong.wang2\}@mail.mcgill.ca,
lijun.sun@mcgill.ca}

\usepackage{bibentry}

\begin{document}

\maketitle

\begin{abstract}
This paper introduces a data-driven time embedding method for modeling long-range seasonal dependencies in spatiotemporal forecasting tasks. The proposed approach employs Dynamic Mode Decomposition (DMD) to extract temporal modes directly from observed data, eliminating the need for explicit timestamps or hand-crafted time features. These temporal modes serve as time representations that can be seamlessly integrated into deep spatiotemporal forecasting models. Unlike conventional embeddings such as time-of-day indicators or sinusoidal functions, our method captures complex multi-scale periodicity through spectral analysis of spatiotemporal data. Extensive experiments on urban mobility, highway traffic, and climate datasets demonstrate that the DMD-based embedding consistently improves long-horizon forecasting accuracy, reduces residual correlation, and enhances temporal generalization. The method is lightweight, model-agnostic, and compatible with any architecture that incorporates time covariates.
\end{abstract}

%

\section{Introduction}

Spatiotemporal forecasting plays a central role in diverse real-world applications, including traffic flow prediction~\citep{vlahogianni2014short}, climate modeling~\citep{climate}, energy demand estimation~\citep{hong2016probabilistic}, and environmental monitoring~\citep{pettorelli2014satellite}. These systems often exhibit strong periodic or seasonal patterns—such as daily commuting cycles, weekly temperature fluctuations, or yearly rainfall changes—that are essential for accurate long-term forecasting and planning. However, capturing such long-range periodic structures remains a persistent challenge for spatiotemporal forecasting~\citep{vlahogianni2014short, karlaftis2011statistical}.

Recent advances in deep learning have enabled powerful spatiotemporal forecasting models by combining neural sequence architectures and spatial representations, such as Recurrent Neural Networks (RNNs)~\citep{lstm, gru}, Temporal Convolutional Networks (TCNs)~\citep{yu2015multi}, Graph Convolutional Networks (GCNs)~\citep{defferrard2016convolutional, kipf2016semi}, and Transformer-based models~\citep{cai2020traffic, grigsby2021long}. While these models effectively capture short-term dynamics, they often struggle to represent global temporal dependencies unless aided by explicit temporal encodings. A common workaround is to include hand-crafted time features (e.g., \textit{time-of-day}, \textit{day-of-week})~\citep{kazemi2019time2vec}, but such features are often domain-specific and insufficient to express complex multi-scale seasonality.

To address this limitation, we propose a data-driven approach to extract and encode periodic structure directly from observations, using \textbf{Dynamic Mode Decomposition (DMD)}~\citep{schmid2022dynamic}. By leveraging Koopman operator theory, DMD decomposes spatiotemporal signals into interpretable oscillatory modes without requiring timestamp metadata. We construct time embeddings by extracting the real and imaginary components of the dominant DMD modes~\citep{brunton2016extracting, jovanovic2014sparsity}, resulting in a compact spectral representation of temporal dynamics. These embeddings can be seamlessly integrated as covariates into any forecasting model to enhance its ability to capture global periodic dependencies.

Our method offers three key advantages:
\begin{enumerate}
    \item \textbf{Interpretable}: The learned embeddings reveal dominant periodic components, providing insight into temporal patterns in the data.
    \item \textbf{Model-agnostic}: The embeddings can be incorporated into any spatiotemporal forecasting architecture that supports time covariates.
    \item \textbf{Domain-agnostic}: The method is purely data-driven and does not rely on calendar time or hand-crafted features, making it broadly applicable across domains.
\end{enumerate}

We evaluate our method on three real-world spatiotemporal datasets: GZ-METRO (urban metro ridership)~\citep{li2017diffusion}, PEMS04 (highway traffic flow)~\citep{song2020spatial}, and Daymet (daily climate temperature)~\citep{thornton2022daymet}. We integrate our DMD-based time embeddings into several deep learning models and demonstrate improved long-term forecasting accuracy, reduced residual correlations, and enhanced temporal generalization across diverse settings.


\begin{table*}[ht]
  \centering
  \caption{Comparison of Time Embedding Methods}
  \begin{tabularx}{\linewidth}{@{}>{\bfseries}l
                                >{\raggedright\arraybackslash}X
                                >{\raggedright\arraybackslash}X
                                >{\raggedright\arraybackslash}X@{}}
    \toprule
    Method                    & Expressiveness                        & Interpretability                            & Generalization                          \\
    \midrule
    Time2Vec
      & \makecell[l]{Learns both periodic\\and non‐periodic patterns}
      & \makecell[l]{Medium: internal weights not\\directly interpretable}
      & \makecell[l]{Strong: end‐to‐end learning,\\no manual design}         \\
    \addlinespace
    Fourier Features
      & \makecell[l]{Fixed‐frequency\\sinusoids}
      & \makecell[l]{High: clear spectral\\interpretation}
      & \makecell[l]{Limited: requires predefined\\frequencies; \\ less robust to non‐stationarity} \\
    \addlinespace
    Handcrafted Features
      & \makecell[l]{Predefined calendars\\and events}
      & \makecell[l]{High: semantically\\intuitive}
      & \makecell[l]{Limited: domain‐specific;\\poor cross‐task transfer}  \\
    \addlinespace
    
    DMD-based Embedding
      & \makecell[l]{Automatically extracts\\multiple periodic modes}
      & \makecell[l]{High: modes correspond\\to spectral components}
      & \makecell[l]{Strong: fully data‐driven,\\domain‐agnostic}            \\
    \bottomrule
  \end{tabularx}
  \label{tab:time-embedding-comparison}
\end{table*}

\section{Related Work}
\label{sec:related}
\vspace{-2pt}
\subsection{Spatiotemporal Forecasting Models}
Existing spatiotemporal forecasting methods can be broadly categorized into three families. 

\textbf{Statistical and linear algebraic models.} Classical statistical models analyze time series in either the time domain (e.g., ARMA, ARIMA, SARIMA~\citep{box2015time}) or frequency domain (e.g., Fourier and wavelet transforms~\citep{cryer2008time}). Matrix factorization techniques have also been applied to low-rank spatiotemporal forecasting and imputation tasks~\citep{yu2016temporal}. 

\textbf{Dynamic systems models.} DMD~\citep{schmid2022dynamic, avila2020data} and Sparse Identification of Nonlinear Dynamics (SINDy)~\citep{brunton2016discovering, champion2019data} aim to recover governing structures from observations using data-driven operator-based formulations.

\textbf{Machine learning models.} Early approaches include tree ensembles and support vector machines~\citep{bishop2006pattern}, while deep learning (DL) models have become the dominant paradigm for capturing complex spatiotemporal dependencies~\citep{lecun2015deep}. Our work contributes to this direction by enhancing DL models with a novel spectral time embedding.
\vspace{-2pt}
\subsection{Deep Learning for Spatiotemporal Forecasting}
DL-based spatiotemporal models combine temporal encoders and spatial graph modules. DCRNN~\citep{li2017diffusion} integrates diffusion convolution into GRUs to model both space and time. STGCN~\citep{yu2017spatio} uses GCNs for spatial structure and CNNs for temporal encoding. ASTGCN~\citep{guo2019attention} adds spatial-temporal attention to the STGCN backbone.

Graph WaveNet~\citep{wu2019graph} introduces an adaptive adjacency matrix and dilated temporal convolutions, offering flexibility in learning the spatial graph structure. STSGCN~\citep{song2020spatial} models temporal adjacency by linking graphs across consecutive time steps. GMAN~\citep{zheng2020gman} adopts dual attention over space and time with embedding fusion. FC-GAGA~\citep{oreshkin2021fc} extends N-BEATS~\citep{oreshkin2019n} with a gated graph layer that learns sparse, non-Markovian structures.

While spatial modules have become increasingly expressive, temporal modeling in these architectures often remains confined to short-range input windows (e.g., one hour), making it difficult to capture long-term seasonal dynamics.
\vspace{-2pt}
\subsection{Time Embedding and Periodic Structure Learning}

Encoding long-term temporal dependency is critical in spatiotemporal forecasting due to prevalent seasonal patterns. The most common practice is to use hand-crafted features such as \textit{time-of-day} or \textit{day-of-week}, which capture fixed daily or weekly periodicities. However, these features are rigid and may not reflect latent periodic structures in the data. Time2Vec~\citep{kazemi2019time2vec} extends sinusoidal positional encoding~\citep{transformer} by learning frequencies and phase shifts, offering a flexible way to encode global time. This method has been widely adopted in forecasting tasks~\citep{grigsby2021long} and is related to Fourier feature mappings~\citep{tancik2020fourier, li2021learnable}. However, recent studies suggest that Fourier features often overfit high-frequency fluctuations and struggle to encode low-frequency seasonality~\citep{tancik2020fourier}.

In contrast, we propose a spectral embedding approach based on DMD. Our method extracts dominant oscillatory modes from observed time series via Koopman operator analysis~\citep{brunton2016extracting, jovanovic2014sparsity}. The real and imaginary components of these modes form a compact and interpretable time representation, which captures complex and data-driven periodic patterns beyond hand-crafted or fixed-frequency alternatives. These embeddings can be seamlessly integrated into any DL forecasting model to enhance long-term temporal modeling.
\vspace{-2pt}
\paragraph{Comparison of Time Embedding Methods.} As Table~\ref{tab:time-embedding-comparison} shows, traditional Fourier and handcrafted time features offer clear interpretability but lack flexibility for complex or non‐stationary signals. Time2Vec can flexibly fit a variety of temporal patterns via end‐to‐end training, yet its internal representations are not readily interpretable. In contrast, our DMD-based embedding automatically identifies and separates dominant spectral modes—providing both high expressiveness and clear modal interpretation—while remaining entirely data‐driven and broadly generalizable.

In the next section, we build on these advantages to introduce our full pipeline for spatiotemporal forecasting.  

\begin{figure*}[htbp]
  \centering
  \includegraphics[width=0.9\textwidth]{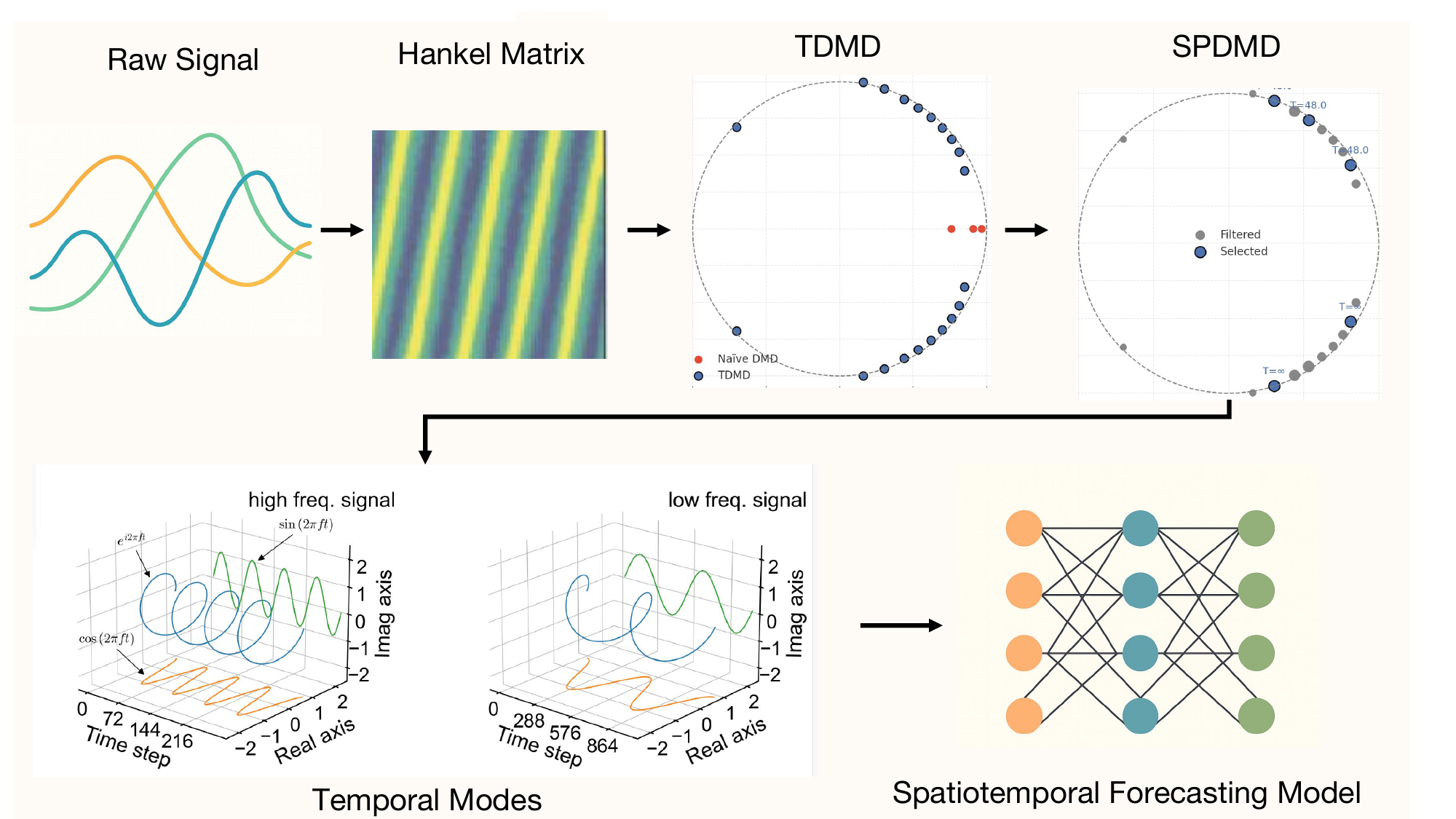}
  \caption{Proposed pipeline. Raw multivariate signals are first converted into a stacked Hankel matrix and then decomposed via Total TDMD. Sparsity-Promoting SPDMD selects the most prominent eigenmodes; their real and imaginary components yield compact cosine–sine time covariates. These covariates are concatenated with other features and fed into any spatiotemporal forecasting model.}
  \label{fig:pipeline_overview}
\end{figure*}
\vspace{-2pt}
\section{Methodology}
\label{sec:method}
In this section, we introduce our method to obtain the spectral properties of spatiotemporal data and the method to encode periodic information in forecasting models.
\subsection{Spatiotemporal Forecasting}

We consider a general spatiotemporal system observed over a network, which can be represented as a directed graph $\mathcal{G}=\left(\mathcal{V}, \mathcal{E}, \mathbf{A}\right)$, where $\mathcal{V}$ is a set of spatial locations or sensing nodes with cardinality $|\mathcal{V}| = N$, $\mathcal{E}$ is a set of edges representing spatial relationships, and $\mathbf{A} \in \mathbb{R}^{N \times N}$ is the weighted adjacency matrix encoding spatial proximity or correlation.

Let $\mathbf{z}_{t} \in \mathbb{R}^{N}$ denote the signal observed at all $N$ locations at time $t$, and let $\mathbf{c}_{t} \in \mathbb{R}^{D}$ represent auxiliary covariates (e.g., temporal features) associated with time $t$. The spatiotemporal forecasting task aims to learn a function $f(\cdot)$ that maps the past $P$ time steps of observations and covariates to the next $Q$ steps of future signals:
\begin{equation}
\label{eqn:forecasting}
    \left[\mathbf{X}_{t-P+1:t}, \mathbf{C}_{t-P+1:t+Q}\right] \xrightarrow{f(\cdot)} \left[\mathbf{Y}_{t+1:t+Q}\right],
\end{equation}
where $\mathbf{X}_{t-P+1:t} = \{\mathbf{z}_{t-P+1}, \dots, \mathbf{z}_{t}\}$ is the historical sequence of observed signals, $\mathbf{C}_{t-P+1:t+Q} = \{\mathbf{c}_{t-P+1}, \dots, \mathbf{c}_{t+Q}\}$ is the sequence of covariates (possibly including future ones), and $\mathbf{Y}_{t+1:t+Q} = \{\mathbf{z}_{t+1}, \dots, \mathbf{z}_{t+Q}\}$ is the target sequence to be predicted.

The function $f(\cdot)$ typically takes the form of a multivariate sequence-to-sequence (Seq2Seq) model and can be parameterized by either classical statistical models or modern deep learning models. In practice, the forecasting horizon $Q$ is often set equal to the historical window length $P$.

\subsection{Spectral Analysis of Spatiotemporal Data with DMD}

To extract periodic structure from spatiotemporal data, we propose a three-stage spectral analysis pipeline based on DMD. The method consists of:
\begin{enumerate}
    \item \textbf{Hankel embedding} to expand the effective spatial dimension of the data (since typically $N \ll T$ in spatiotemporal forecasting)~\citep{champion2019discovery, kamb2020time, wang2022extracting};
    \item \textbf{Total DMD (TDMD)} for unbiased, noise-aware spectral decomposition~\citep{hemati2017biasing};
    \item \textbf{Sparsity-Promoting DMD (SPDMD)} to select dominant dynamic modes~\citep{jovanovic2014sparsity}.
\end{enumerate}

The complete processing chain is summarised in Figure \ref{fig:pipeline_overview}. Each stage will be detailed next. Our goal is to obtain a compact and interpretable set of oscillatory components whose temporal evolution will serve as covariates in downstream forecasting models.

Let $\mathbf{z}_t \in \mathbb{R}^N$ denote the observed system state at time step $t$, across $N$ spatial nodes. The dynamics of the system can be modeled as a nonlinear discrete-time system:
\begin{equation}
\mathbf{z}_{t+1} = f(\mathbf{z}_t),
\end{equation}
where $f$ is an unknown nonlinear function. To linearize the system, we follow the Koopman operator framework~\citep{schmid2022dynamic}, which lifts the dynamics into a higher-dimensional observable space via a function $\phi(\cdot)$ such that:
\begin{equation}
\phi(\mathbf{z}_{t+1}) = \mathcal{K} \phi(\mathbf{z}_t),
\end{equation}
where $\mathcal{K}$ is a linear (but infinite-dimensional) Koopman operator. Its spectral decomposition yields:
\begin{equation}
\mathcal{K} \boldsymbol{\Psi} = \boldsymbol{\Psi} \boldsymbol{\Lambda},
\end{equation}
where $\boldsymbol{\Psi} \in \mathbb{C}^{d \times r}$ contains Koopman eigenfunctions and $\boldsymbol{\Lambda} = \mathrm{diag}(\lambda_1, \dots, \lambda_r)$ is a diagonal matrix of corresponding eigenvalues. Assuming the initial observable satisfies $\phi(\mathbf{z}_0) = \boldsymbol{\Psi} \mathbf{a}$, the time evolution becomes:
\begin{equation}
\phi(\mathbf{z}_t) = \mathcal{K}^t \phi(\mathbf{z}_0) = \boldsymbol{\Psi} \boldsymbol{\Lambda}^t \mathbf{a}.
\end{equation}

Let $\boldsymbol{\phi}_t = \phi(\mathbf{z}_t)$ denote the observable at time $t$. We stack the raw input signals over time into the data matrix $\mathbf{D} = [\mathbf{z}_0, \mathbf{z}_1, \dots, \mathbf{z}_{T-1}] \in \mathbb{R}^{N \times T}$. Its Koopman-based approximation becomes:
\begin{equation}
\label{eq:dmd}
\hat{\mathbf{D}} = \boldsymbol{\Psi} \cdot \mathrm{diag}(\mathbf{a}) \cdot \mathbf{C},
\end{equation}
where $\mathbf{C} \in \mathbb{C}^{r \times T}$ is a Vandermonde matrix with:
\[
\mathbf{C}_{ij} = \lambda_i^{j-1}, \quad i = 1,\dots,r, \; j = 1,\dots,T.
\]

\begin{figure}[ht]
  \centering
  \includegraphics[width=0.8\linewidth]{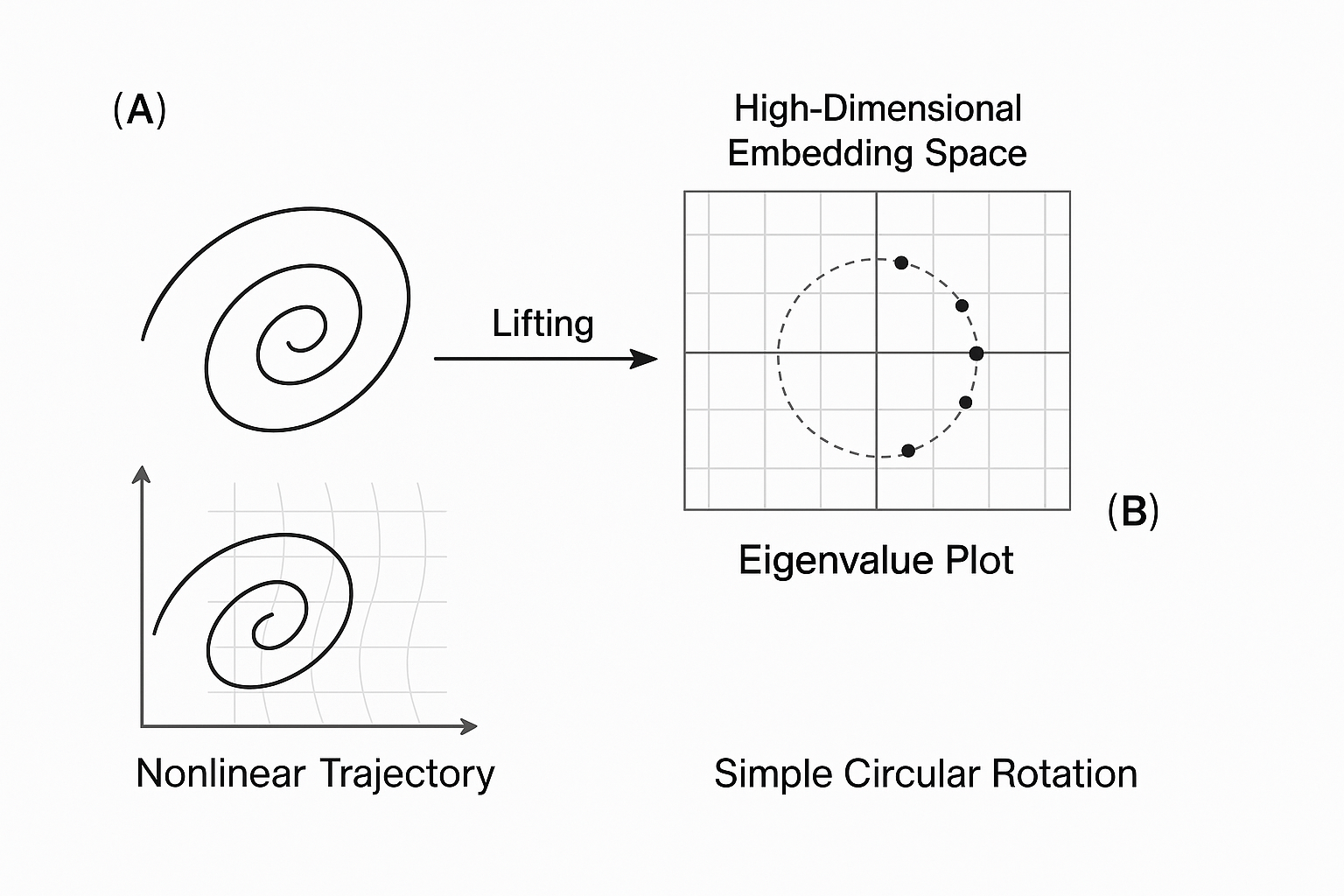}
  \caption{Koopman operator intuition.
    (A) A nonlinear spiral trajectory is lifted into a higher-dimensional observable space.
    (B) In that space the dynamics reduce to a simple rotation; the eigenvalues of the Koopman operator lie on (or near) the unit circle, each dot representing an oscillatory mode whose frequency is given by its angle.}
  \label{fig:koopman_schematic}
\end{figure}

Figure \ref{fig:koopman_schematic} illustrates this idea: a nonlinear spiral trajectory (left) is “lifted” into a high-dimensional space where the Koopman operator acts as a simple rotation; its eigenvalues appear as points on/near the unit circle.

To encourage interpretability, we apply SPDMD~\citep{jovanovic2014sparsity} to prune to the $r$ dominant modes, yielding:
\begin{equation}
\label{eq:rankr}
\hat{\mathbf{D}} \approx \bar{\boldsymbol{\Psi}} \cdot \mathrm{diag}(\bar{\mathbf{a}}) \cdot \bar{\mathbf{C}},
\end{equation}
where $\bar{\boldsymbol{\Psi}} \in \mathbb{C}^{N \times r}$ contains selected spatial modes, $\bar{\mathbf{a}} \in \mathbb{C}^r$ are their amplitudes, and $\bar{\mathbf{C}} \in \mathbb{C}^{r \times T}$ captures their temporal evolution.

To improve the performance of DMD on high-dimensional data, we first enhance the data using a circulant Hankel embedding~\citep{schmid2022dynamic, wang2023anti}, which converts $\mathbf{D}$ into a tall matrix:
\begin{equation}
\mathbf{H} =
\begin{bmatrix}
\mathbf{z}_0 & \mathbf{z}_1 & \cdots & \mathbf{z}_{T-1} \\
\mathbf{z}_1 & \mathbf{z}_2 & \cdots & \mathbf{z}_0 \\
\vdots & \vdots & \ddots & \vdots \\
\mathbf{z}_{\tau-1} & \mathbf{z}_\tau & \cdots & \mathbf{z}_{\tau-2}
\end{bmatrix} \in \mathbb{R}^{N\tau \times T}.
\label{eq:hankel}
\end{equation}

Since directly computing the SVD of $\mathbf{H} \mathbf{H}^\top \in \mathbb{R}^{N\tau \times N\tau}$ is computationally expensive for large $\tau$, we apply the \textit{method of snapshots}~\citep{brunton2022data}, which leverages the symmetry of the inner product space to reduce the computation to a smaller Gram matrix. Specifically, we perform the truncated SVD on the smaller matrix $\mathbf{H}^\top \mathbf{H} \in \mathbb{R}^{T \times T}$:
\begin{equation}
    \mathbf{H}^\top \mathbf{H} = \tilde{\mathbf{V}} \tilde{\mathbf{\Sigma}}^2 \tilde{\mathbf{V}}^\top,
\label{eq:svd}
\end{equation}

where $\tilde{\mathbf{V}} \in \mathbb{R}^{T \times r}$ contains the top-$r$ right singular vectors; $\tilde{\mathbf{\Sigma}} \in \mathbb{R}^{r \times r}$ is the diagonal matrix of singular values.

Then, we recover the corresponding left singular vectors using:
\begin{equation}
    \tilde{\mathbf{U}} = \mathbf{H} \tilde{\mathbf{V}} \tilde{\mathbf{\Sigma}}^{-1},
    \label{eq:recover}
\end{equation}
with $\tilde{\mathbf{U}} \in \mathbb{R}^{N\tau \times r}$ forming an orthonormal basis for the dominant subspace of $\mathbf{H}$.

Let $\mathbf{H}' \in \mathbb{R}^{N\tau \times T}$ denote the time-shifted version of $\mathbf{H}$, i.e., containing the next-step snapshots. Using $\tilde{\mathbf{U}}$ and $\tilde{\mathbf{V}}$, we construct the reduced Koopman operator:
\begin{equation}
\mathbf{A}_{\text{DMD}} = \tilde{\mathbf{U}}^\top \mathbf{H}' \tilde{\mathbf{V}} \tilde{\mathbf{\Sigma}}^{-1},
\label{eq:form}
\end{equation}
which approximates the linear transition matrix in the low-rank temporal subspace. The eigenvalues $\{\lambda_1, \dots, \lambda_r\}$ of $\mathbf{A}_{\text{DMD}}$ characterize the oscillatory behavior of the system (magnitude $\rightarrow$ growth/decay, imaginary part $\rightarrow$ frequency)~\citep{schmid2022dynamic}.


From these eigenvalues, we construct the temporal dynamics matrix $\bar{\mathbf{C}} \in \mathbb{C}^{r \times T}$ as a Vandermonde matrix:
\begin{equation}
\bar{\mathbf{C}} = 
\begin{bmatrix}
1 & \lambda_1 & \lambda_1^2 & \cdots & \lambda_1^{T-1} \\
1 & \lambda_2 & \lambda_2^2 & \cdots & \lambda_2^{T-1} \\
\vdots & \vdots & \vdots & \ddots & \vdots \\
1 & \lambda_r & \lambda_r^2 & \cdots & \lambda_r^{T-1}
\end{bmatrix}.
\end{equation}
Each row in $\bar{\mathbf{C}}$ defines the temporal profile of a dominant DMD mode. These complex-valued signals form the basis of our time embedding.
\vspace{-2pt}
\paragraph{Final Output: Time Embedding for Forecasting.}
At each time step $t$, we extract the real and imaginary parts of $\bar{\mathbf{C}}[:, t]$ to construct a $2r$-dimensional time covariate:
\begin{equation}
    \mathbf{c}_t^{(\text{DMD})} = 
\begin{bmatrix}
\mathrm{Re}(\bar{\mathbf{C}}[:, t]) \\
\mathrm{Im}(\bar{\mathbf{C}}[:, t])
\end{bmatrix} \in \mathbb{R}^{2r}.
\end{equation}
This embedding can be appended to any spatiotemporal forecasting model. If the original model input has shape $(N, T, m)$, representing spatial dimension, temporal steps, and feature channels respectively, then our method expands it to shape $(N, T, m + 2r)$ by concatenating DMD-based time embeddings at each time step. 
\vspace{-2pt}
\paragraph{Remark.}
Each DMD eigenvalue can be written as a complex exponential form \( \lambda_k = e^{\mu_k + i\omega_k} \), where \( \mu_k \) denotes the growth or decay rate, and \( \omega_k \) denotes the oscillation frequency. 
In our construction of time covariates, we exclusively use the frequency components \( \omega_k \) to generate sinusoidal embeddings:
\[
\mathbf{c}_t^{(\text{DMD})} = [\cos(\omega_1 t), \ldots, \cos(\omega_K t), \sin(\omega_1 t), \ldots, \sin(\omega_K t)].
\]
The amplitude dynamics \( e^{\mu_k t} \) are discarded to avoid potential numerical instability and to ensure the covariates remain interpretable and stationary. 
This design decouples trend-like dynamics from periodic components, allowing the main model to focus on learning non-periodic residual structures.

\begin{algorithm}[ht]
\caption{DMD-based Time Covariate Construction}
\label{alg:dmd_embedding}
\begin{algorithmic}[1]
\REQUIRE Raw signals $\{\mathbf{z}_t\}_{t=0}^{T-1}$, Hankel window $\tau$, mode count $r$
\ENSURE Time covariates $\{\mathbf{c}_t^{(\text{DMD})}\}_{t=0}^{T-1}$
\STATE Build stacked Hankel matrix $\mathbf{H}$ (Eq.~\ref{eq:hankel})
\STATE Compute SVD: 
  $\mathbf{H}^\top \mathbf{H}
    = \widetilde{\mathbf{V}}
      \,\widetilde{\mathbf{\Sigma}}^2\,
      \widetilde{\mathbf{V}}^\top$
      (Eq.~\ref{eq:svd})
\STATE Recover 
  $\widetilde{\mathbf{U}}
    = \mathbf{H}\,
      \widetilde{\mathbf{V}}\,
      \widetilde{\mathbf{\Sigma}}^{-1}$
      (Eq.~\ref{eq:recover})
\STATE Form reduced operator: 
  $\mathbf{A}_{\text{DMD}}
    = \widetilde{\mathbf{U}}^\top
      \mathbf{H}'\,
      \widetilde{\mathbf{V}}\,
      \widetilde{\mathbf{\Sigma}}^{-1}$
      (Eq.~\ref{eq:form})
\STATE Compute eigenpairs 
  $\{(\lambda_k,\psi_k)\}_{k=1}^r$ of $\mathbf{A}_{\text{DMD}}$
\FOR{$t=0$ \TO $T-1$}
  \STATE $\mathbf{c}_t^{(\text{DMD})} = [\cos(\omega_1 t),\dots,\sin(\omega_r t)]$
    with $\omega_k = \arg(\lambda_k)$
\ENDFOR
\RETURN $\{\mathbf{c}_t^{(\text{DMD})}\}$
\end{algorithmic}
\end{algorithm}

\section{Experiments}
\label{sec:experiments}
To evaluate the effectiveness of the proposed method, we carried out experiments on three spatiotemporal datasets: GZ-METRO, PEMS04, and Daymet. GZ-METRO records 15-min ridership based on smart card tap-in data at 159 metro stations for 3 months in Guangzhou, China. PEMS04 is a highway traffic flow datasets used in \citep{song2020spatial}. The dataset records 5-min aggregated traffic flow from the Cal-trans Performance Measurement System (PeMS) \citep{chen2001freeway}. Daymet is a dataset that provides the daily maximum temperature recordings for Continental North America from 2010 to 2021 \citep{thornton2022daymet}. The summary of datasets can be found in Table~\ref{tab:datasets}. For GZ-METRO and Daymet, 70\% of the data were used for training, 20\% for testing, and the remaining 10\% for validation. For PEMS04, the split ratio is 6:2:2 as used in the original paper \citep{song2020spatial}. All datasets were applied z-score normalization with statistics obtained from the training set. We selected several spatiotemporal forecasting frameworks as the base models:

\begin{itemize}
    \item FC-LSTM~\citep{sutskever2014sequence}: A Sequence-to-Sequence LSTM model with fully-connected LSTM layers in both encoder and decoder.
    \item DCRNN~\citep{li2017diffusion}: Diffusion convolution recurrent neural network, which wraps diffusion convolution operation into recurrent neural networks to achieve spatiotemporal forecasting.
    \item Graph WaveNet~\citep{wu2019graph}: A spatiotemporal forecasting model that combines dilated 1D convolution for modeling temporal dynamics and graph convolution for modeling spatial dynamics.
    \item AGCRN~\citep{bai2020adaptive}: Similar to Graph WaveNet which uses an adaptive adjacency matrix, the model uses GRU to model temporal dependency.
\end{itemize}

\begin{table}[!h]
\caption{Dataset Description\label{tab:datasets}}
\centering
\begin{tabular}{c|c|c}
\toprule
Datasets & Number of nodes & Time range\\
\midrule
\midrule
GZ-METRO & 159 & 7/1/2017 - 9/29/2017 \\
PEMS04 & 307 & 1/1/2018 - 2/28/2018  \\
Daymet & 628 & 1/1/2010 - 12/31/2021  \\
\bottomrule
\end{tabular}
\end{table}

We implemented these models using the original source code (or their PyTorch version). All models use 12 steps of historical observations ($P=12$) to predict 12 steps of future values ($Q=12$). The evaluation metrics are mean absolute error (MAE) and root mean squared error (RMSE), where missing values are excluded:
\begin{equation}
\text{MAE}=\frac{1}{N} \sum_{i=1}^{N} \left|z_{i} - \hat{z}_{i}\right|,
\end{equation}
\begin{equation}
\text{RMSE}=\sqrt{\frac{1}{N} \sum_{i=1}^{N} \left(z_{i} - \hat{z}_{i}\right)^2}.
\end{equation}

Early stopping was applied to prevent over-fitting when the validation loss keeps growing for more than 30 epochs. The learning rate was set to 0.001 with a decay rate of 0.0001. The Adam algorithm was used for optimization. Our experiments were conducted under a computer environment with one Intel(R) Xeon(R) CPU E5-2698 v4 @ 2.20GHz and four NVIDIA Tesla V100 GPU. All results were computed based on the average of three runs of training. 


\vspace{-2pt}
\subsection{Quantitative Study}
\begin{figure}[!t]
\centering
\includegraphics[width=0.95\linewidth]{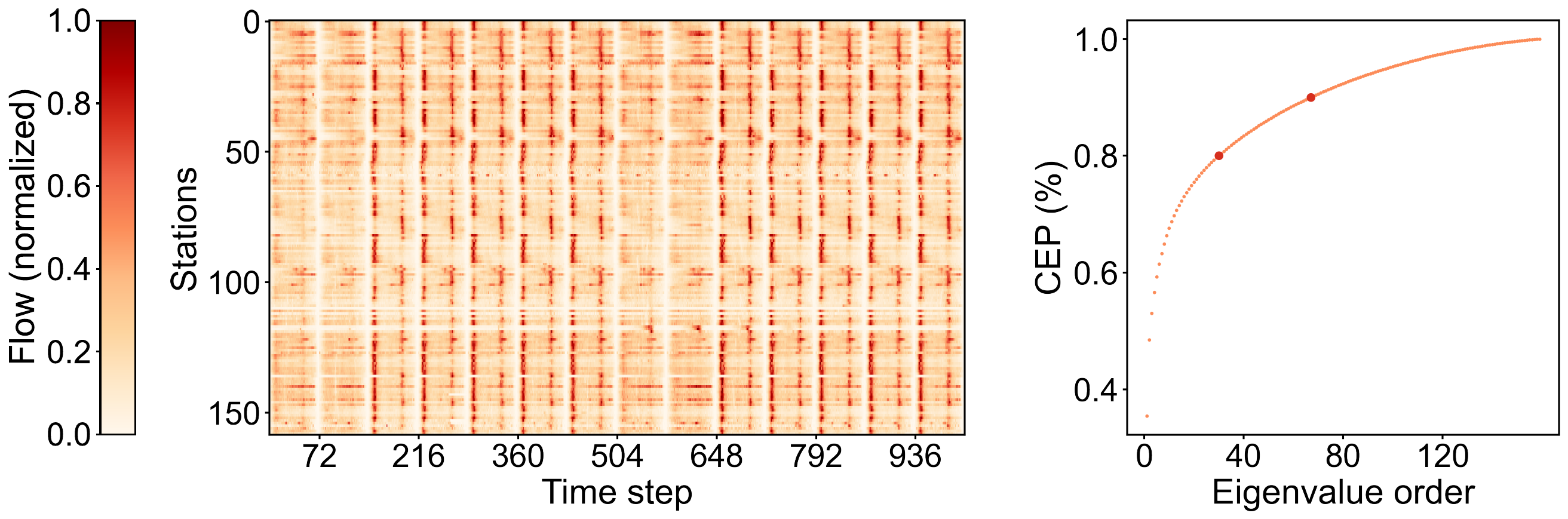}
\caption{Left: Normalized ridership at all stations of the Guangzhou Metro. Right: Cumulative eigenvalue percentage (CEP) calculated by SVD.}
\label{fig:gz_example}
\end{figure}

We begin our quantitative analysis by examining the temporal and spectral characteristics of the GZ-METRO dataset. Figure~\ref{fig:gz_example} (left) visualizes the ridership flow over two weeks for all metro stations. The repeated daily and weekly patterns suggest the presence of strong periodic signals in both time and space. To further quantify this structure, we perform singular value decomposition (SVD) and compute the cumulative eigenvalue percentage (CEP), as shown in Figure~\ref{fig:gz_example} (right). We observe that more than 90\% of the data variance can be captured using only a small number of singular components, confirming the low-rank nature of the data and motivating the use of spectral methods such as DMD to extract dominant temporal dynamics.

Table~\ref{tab:benchmarks} summarizes the performance of baseline forecasting models on the three datasets, with and without our DMD-based time embedding. Across all datasets and forecasting horizons (3, 6, and 12 steps), we observe consistent improvements when incorporating the proposed time covariates.

On GZ-METRO, the gain is particularly significant at longer horizons (e.g., 12-step ahead), where capturing long-term periodic patterns becomes more important. For instance, Graph WaveNet improves from 93.62 (RMSE) to 87.79 with our embedding, while DCRNN drops from 105.48 to 91.98. This highlights the benefit of DMD-based features in modeling structured temporal dependencies in metro ridership data, which has high regularity and low noise due to 15-minute aggregation.

On PEMS04, which has higher temporal resolution (5-minute) and shorter-range patterns, the performance improvement is less pronounced but still present. For example, AGCRN’s 12-step RMSE improves from 33.5 to 33.16, and Graph WaveNet improves from 33.28 to 32.43. The smaller margin may be attributed to the noisier nature of traffic data and the relatively short forecasting horizon (1 hour), which limits the benefit of global periodic signals.

For the Daymet dataset, which tracks long-term daily temperature patterns, our method also leads to consistent performance improvements. Graph WaveNet’s 12-step RMSE is reduced from 5.7 to 5.27, and AGCRN improves from 5.37 to 5.16. These results support that our time embedding is generalizable across domains and not restricted to transportation data.

Overall, the results suggest that our DMD-based embedding offers a lightweight and effective way to inject long-term periodic structure into spatiotemporal forecasting models, improving both short-term accuracy and long-term generalization. The extracted complex dynamics capture multiple time scales, including daily and weekly cycles, which are otherwise hard to model with conventional hand-crafted or learnable time features alone.

\begin{table}[!t]
\caption{Performance comparison of models with/without time embedding\label{tab:benchmarks}}
\centering
\resizebox{\linewidth}{!}{%
\begin{tabular}{ccccccccc}
\toprule
\multirow{2}{*}{Data} & \multicolumn{2}{c}{\multirow{2}{*}{Model}} & \multicolumn{2}{c}{3-step} & \multicolumn{2}{c}{6-step} & \multicolumn{2}{c}{12-step} \\
\cmidrule(lr){4-9}
&   &  & \multicolumn{1}{c}{MAE} & \multicolumn{1}{c}{RMSE} & \multicolumn{1}{c}{MAE} & \multicolumn{1}{c}{RMSE} & \multicolumn{1}{c}{MAE} & \multicolumn{1}{c}{RMSE} \\
\midrule
\midrule
\multirow{8}{*}{\rotatebox[origin=c]{90}{GZ-METRO}} 
& \multirow{2}{*}{FC-LSTM} & w/o & 48.05 & 92.33 & 50.44 & 97.43 & 53.38 & 103.57\\
&  & w/  & \textbf{47.4} & \textbf{90.3} & \textbf{49.02} & \textbf{94.13} & \textbf{51.18} & \textbf{99.66} \\
\cmidrule(lr){3-9}
& \multirow{2}{*}{DCRNN} & w/o & 43.39 & 83.75 & 49.08 & 92.35 & 55.43 & 105.48\\
& & w/ & \textbf{42.32} & \textbf{80.29} & \textbf{45.46} & \textbf{84.71} & \textbf{50.07} & \textbf{91.98} \\
\cmidrule(lr){3-9}
& \multirow{2}{*}{AGCRN} & w/o & 42.66 & 81.5 & 47.12 & 91.14 & 51.13 & 97.44 \\
& & w/ & \textbf{40.7} & \textbf{77.61} & \textbf{44.35} & \textbf{84.08} & \textbf{48.77} & \textbf{91.46} \\
\cmidrule(lr){3-9}
& \multirow{2}{*}{Graph WaveNet} & w/o & 41.26 & 76.33 & 44.66 & 82.82 & 50.05 & 93.62 \\
& & w/ & \textbf{39.47} & \textbf{73.21} & \textbf{42.87} & \textbf{79.99} & \textbf{47.53} & \textbf{87.79} \\
\midrule
\multirow{8}{*}{\rotatebox[origin=c]{90}{PEMS04}}
 & \multirow{2}{*}{FC-LSTM} & w/o & 22.36 & 36.44 & 22.52 & 36.57 & 22.96 & 36.99\\
 & & w/ & \textbf{22.11} & \textbf{36.21} & \textbf{22.17} & \textbf{36.28} & \textbf{22.41} & \textbf{36.51} \\
\cmidrule(lr){3-9}
 & \multirow{2}{*}{DCRNN} & w/o & 19.89 & 31.29 & 21.95 & 34.46 & 26.1 & 40.81\\
 & & w/ & \textbf{19.01} & \textbf{30.32} & \textbf{20.2} & \textbf{32.34} & \textbf{22.14} & \textbf{35.42} \\
\cmidrule(lr){3-9}
& \multirow{2}{*}{AGCRN} & w/o & 18.65 & \textbf{29.83} & 19.56 & 31.41 & 21.05 & 33.5 \\
& & w/ & \textbf{18.48} & 29.88 & \textbf{19.24} & \textbf{31.34} & \textbf{20.44} & \textbf{33.16} \\
\cmidrule(lr){3-9}
 & \multirow{2}{*}{Graph WaveNet} & w/o & 18.2 & 29.1 & 19.35 & 30.74 & 21.32 & 33.28 \\
 & & w/ & \textbf{18.08} & \textbf{29.03} & \textbf{18.89} & \textbf{30.45} & \textbf{20.23} & \textbf{32.43} \\
\midrule
\multirow{6}{*}{\rotatebox[origin=c]{90}{Daymet}}
 & \multirow{2}{*}{FC-LSTM} & w/o & \textbf{3.89} & \textbf{5.11} & 4.05 & 5.3 & 4.15 & 5.45 \\
 & & w/ & 3.9 & 5.13 & \textbf{4.03} & \textbf{5.29} & \textbf{4.09} & \textbf{5.39} \\
 \cmidrule(lr){3-9}
& \multirow{2}{*}{AGCRN} & w/o & 3.71 & 4.85 & 3.95 & 5.17 & 4.09 & 5.37 \\
& & w/ & \textbf{3.61} & \textbf{4.75} & \textbf{3.86} & \textbf{5.07} & \textbf{3.92} & \textbf{5.16} \\
\cmidrule(lr){3-9}
 & \multirow{2}{*}{Graph WaveNet} & w/o & 3.79 & 4.96 & 4.07 & 5.33 & 4.31 & 5.7 \\
 & & w/ & \textbf{3.76} & \textbf{4.92} & \textbf{3.95} & \textbf{5.17} & \textbf{4.02} & \textbf{5.27} \\
\bottomrule
\end{tabular}}
\end{table}

\begin{table}[!t]
\caption{Performance comparison of different time embedding methods using Graph WaveNet\label{tab:compare_with_t2v}}
\centering
\resizebox{\linewidth}{!}{%
\begin{tabular}{cccccccc}
\toprule
\multirow{2}{*}{Data} & \multirow{2}{*}{Method} & \multicolumn{2}{c}{3-step} & \multicolumn{2}{c}{6-step} & \multicolumn{2}{c}{12-step} \\
\cmidrule(lr){3-8}
&   & MAE & RMSE & MAE & RMSE & MAE & RMSE \\
\midrule
GZ-METRO & D         & 41.48 & 77.23 & 44.89 & 83.64 & 48.72 & 90.62 \\
         & DW        & 43.58 & 83.75 & 47.02 & 91.46 & 50.62 & 96.00 \\
         & Time2Vec  & 40.17 & 74.68 & 44.05 & 81.83 & 48.40 & 91.09 \\
         & Ours      & \textbf{39.47} & \textbf{73.21} & \textbf{42.87} & \textbf{79.99} & \textbf{47.53} & \textbf{87.79} \\
\midrule
PEMS04   & D         & 18.32 & 29.28 & 19.53 & 31.03 & 21.42 & 33.62 \\
         & DW        & 18.13 & 29.11 & 19.16 & 30.64 & 20.73 & 32.86 \\
         & Time2Vec  & 18.26 & 29.21 & 19.48 & 30.93 & 21.47 & 33.62 \\
         & Ours      & \textbf{18.08} & \textbf{29.03} & \textbf{18.89} & \textbf{30.45} & \textbf{20.23} & \textbf{32.43} \\
\midrule
Daymet   & Time2Vec  & \textbf{3.75} & \textbf{4.91} & 4.02 & 5.26 & 4.23 & 5.58 \\
         & Ours      & 3.76 & 4.92 & \textbf{3.95} & \textbf{5.17} & \textbf{4.02} & \textbf{5.27} \\
\bottomrule
\end{tabular}}
\end{table}

\vspace{-2pt}
\subsection{Qualitative Study}

\begin{figure*}[!t]
  \centering
  \begin{subfigure}[t]{0.48\textwidth}
    \centering
    \includegraphics[width=\textwidth]{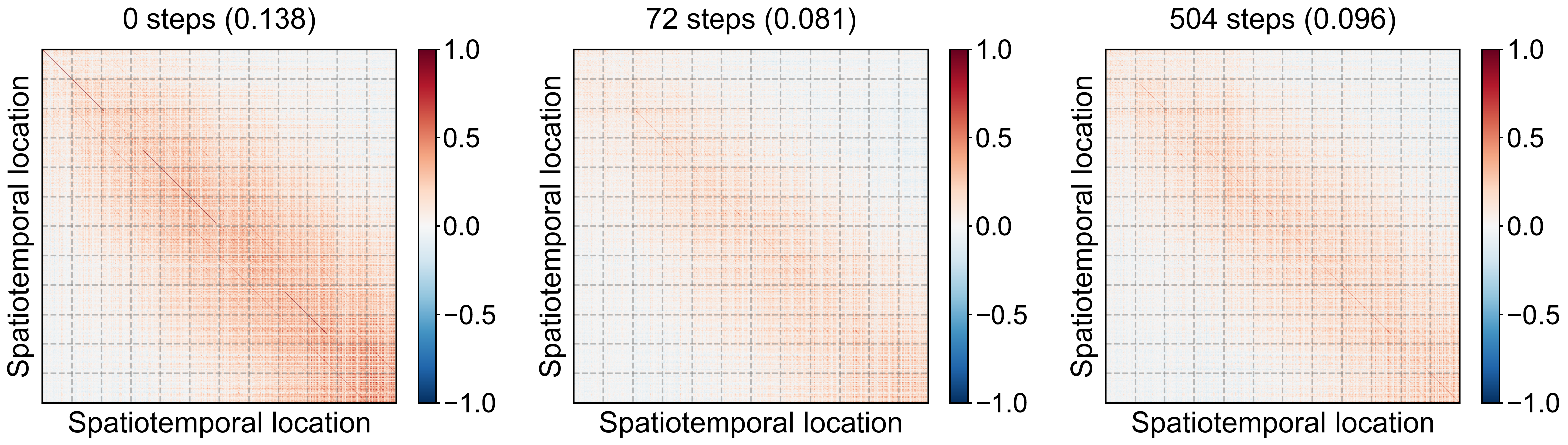}
    \caption{Without time embedding (Graph WaveNet only).}
    \label{fig:res_corr_before}
  \end{subfigure}
  \hfill
  \begin{subfigure}[t]{0.48\textwidth}
    \centering
    \includegraphics[width=\textwidth]{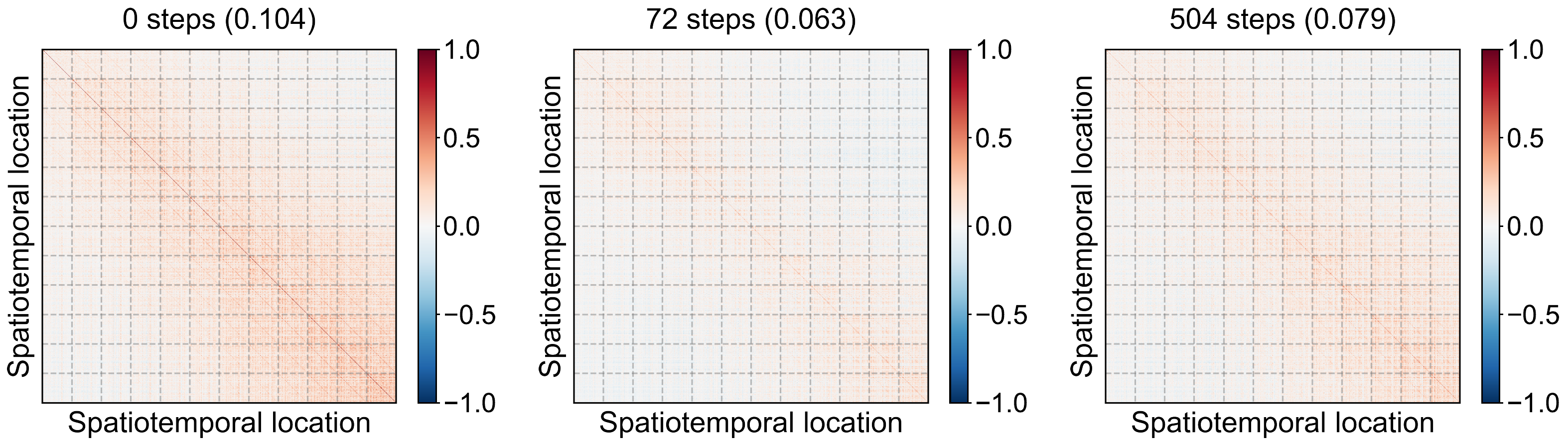}
    \caption{With our method (Graph WaveNet + DMD).}
    \label{fig:res_corr_after}
  \end{subfigure}

  \begin{subfigure}[t]{0.48\textwidth}
    \centering
    \includegraphics[width=\textwidth]{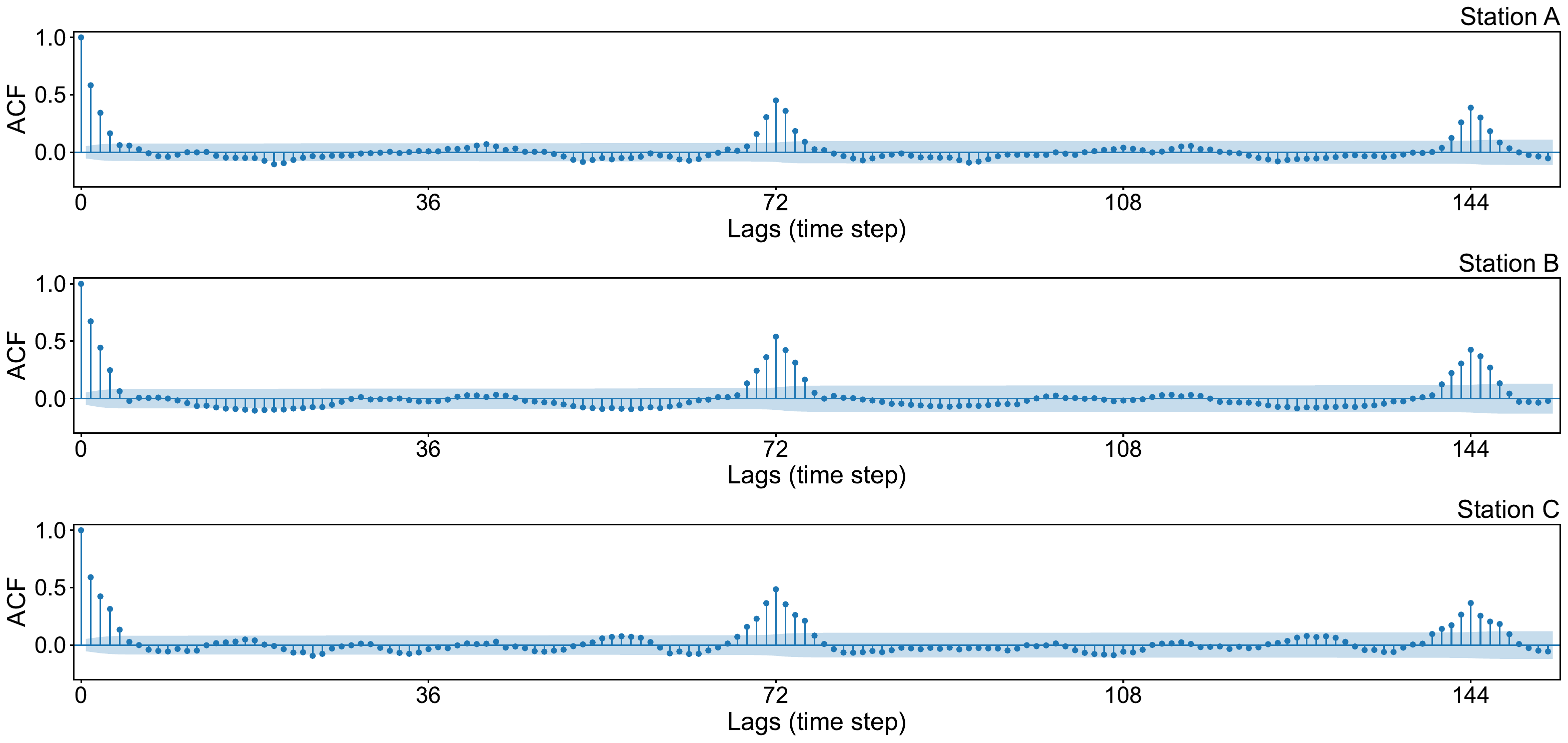}
    \caption{ACF without time embedding.}
    \label{fig:gz_acf}
  \end{subfigure}
  \hfill
  \begin{subfigure}[t]{0.48\textwidth}
    \centering
    \includegraphics[width=\textwidth]{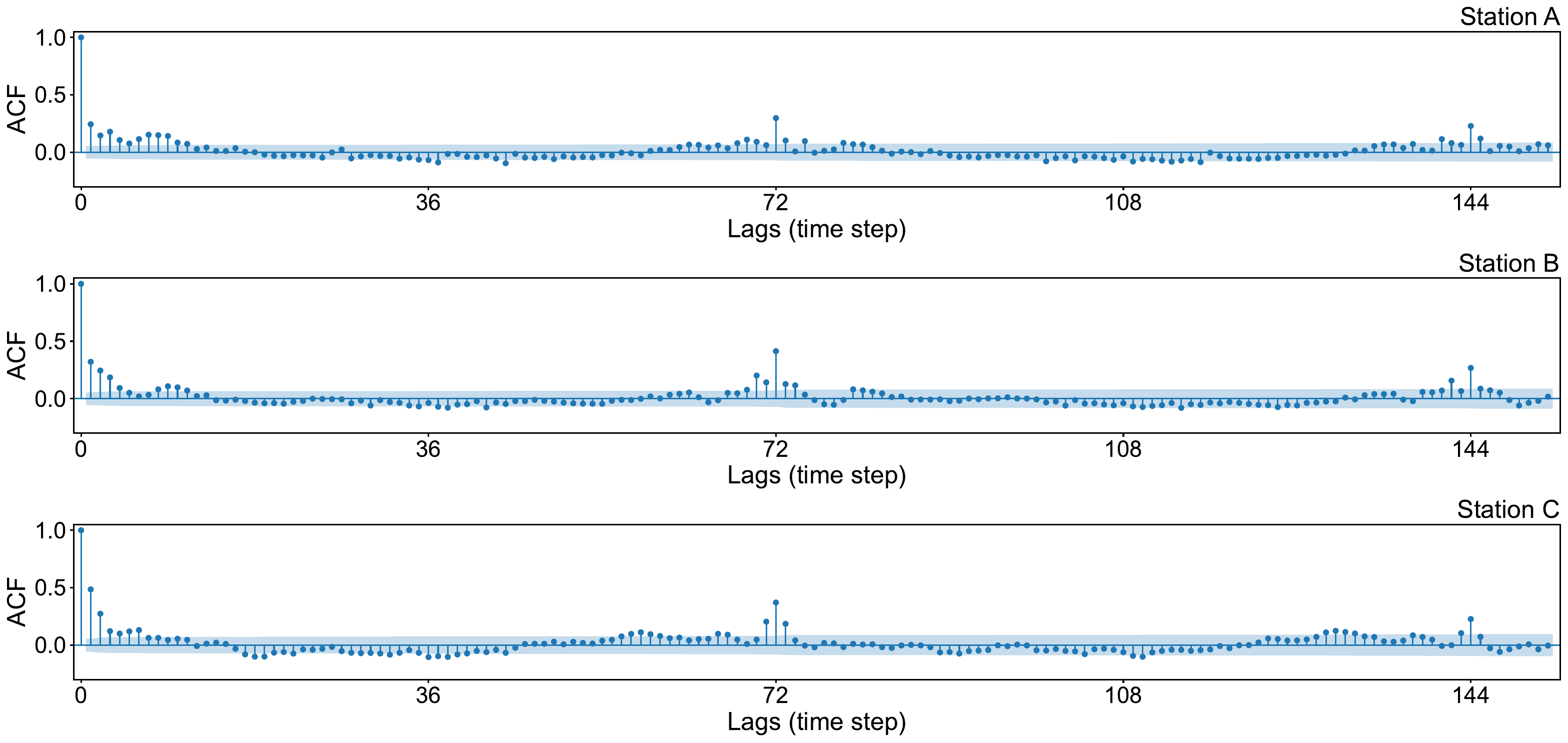}
    \caption{ACF with our method.}
    \label{fig:gz_te_acf}
  \end{subfigure}

  \caption{%
    \textbf{(a)--(b)} Residual correlation matrices of GZ-METRO at lags $S=0,72,504$, 
      without and with our DMD-based embedding. 
    \textbf{(c)--(d)} Autocorrelation functions (ACF) of 12-step prediction residuals 
      from three sensors, showing how our embedding reduces periodic error peaks.
  }
  \label{fig:combined_res_acf}
\end{figure*}

We conduct qualitative analysis of our method using Graph WaveNet as the backbone model and GZ-METRO as the evaluation dataset.
\vspace{-2pt}
\subsubsection{Effectiveness of DMD-based time embedding}

To evaluate the effectiveness of our DMD-based time embedding, we compare it with several alternatives for encoding periodic patterns: 
(i) time of day (D), 
(ii) time of day, day of week (DW), and 
(iii) the learnable sinusoidal embedding method Time2Vec~\citep{kazemi2019time2vec}.

Time of day is encoded as a continuous real-valued feature, whereas day of week is represented using one-hot encoding. To ensure fairness, we align the embedding dimensions of our method and Time2Vec with that of the DW method. For example, if DW uses an embedding dimension of 8, we set the number of modes in our method to $r=4$. As shown in Table~\ref{tab:compare_with_t2v}, our method consistently outperforms all baselines across multiple datasets (GZ-METRO, PEMS04, Daymet) and prediction horizons (3-step, 6-step, 12-step). Notably, in GZ-METRO and PEMS04, our method achieves the lowest RMSE and MAE values in all settings. For example, at 12-step forecasting on GZ-METRO, our method achieves an RMSE of 87.79 compared to 91.09 from Time2Vec and 96.00 from discrete encoding.


Our observations align with the findings of~\citep{tancik2020fourier}, which showed that learned Fourier-based time features tend to concentrate on high-frequency components and struggle to capture long-range periodicity. We further inspected the learned frequencies of Time2Vec and found that they mostly correspond to short-term fluctuations with periods between 5 and 15 time steps (i.e., 25–75 minutes), indicating a failure to model daily or weekly cycles. By contrast, our DMD-based embedding captures multiple frequencies of varying scales, including long-term modes.

\vspace{-2pt}
\subsubsection{Residual correlation and temporal dependency analysis}

To evaluate whether our method effectively reduces systematic errors and better captures long-term temporal dependencies, we conduct two types of residual analysis on the GZ-METRO dataset using Graph WaveNet with and without our DMD-based time embedding.

Figure~\ref{fig:res_corr_before} and Figure~\ref{fig:res_corr_after} show the residual correlation matrices computed from the 12-step-ahead prediction errors at three time lags: $S = 0$, $72$, and $504$ steps, corresponding to contemporaneous, daily, and weekly intervals. Each panel visualizes the correlations between residual vectors $\boldsymbol{\eta}_t$ and $\boldsymbol{\eta}_{t-S}$ over all spatiotemporal positions. Without our method, we observe strong overall correlations at $S=0$ (0.138), and noticeable dependencies at $S=72$ (0.081) and $S=504$ (0.096). In contrast, our method reduces these values to 0.104, 0.063, and 0.079, respectively. These reductions indicate that our model effectively captures structured periodic patterns and reduces error autocorrelation across periods.

To further validate this observation, we analyze the autocorrelation function (ACF) of the prediction residuals from three representative stations (A, B, and C). Figure~\ref{fig:gz_acf} shows the ACF plots of the original Graph WaveNet model. We observe clear peaks around lag 72 and 144, corresponding to daily and bi-daily periodicity, suggesting that the model fails to fully capture these regular patterns, which manifest as cyclic structure in the residuals. In comparison, Figure~\ref{fig:gz_te_acf} illustrates the ACF of the residuals using Graph WaveNet augmented with our time embedding. The periodic peaks at lag 72 and 144 are substantially diminished across all three stations. This indicates that the DMD-based time features help the model internalize the dominant periodic components in the data, leading to less temporally correlated residuals and improved forecasting generalization.

\vspace{-2pt}
\section{Conclusion}
\label{sec:conclusion}
We propose a data-driven time embedding method based on DMD to enhance spatiotemporal forecasting. By extracting dominant temporal modes from the data and representing them through complex-valued dynamic patterns, our method captures long-term periodic structures and injects them as interpretable time covariates into forecasting models. Empirical results across multiple datasets show that our approach consistently improves long-horizon prediction accuracy and reduces residual autocorrelation, especially in datasets with strong seasonal patterns like GZ-METRO. The method is model-agnostic and can be easily integrated into existing architectures with time feature input.

However, on noisy datasets with weak or irregular periodicity—such as those with short aggregation intervals—its benefits are less pronounced. Future work will focus on adapting the method to handle such non-stationary settings and extending it to architectures with different input paradigms, such as Transformers \citep{transformer} where time embeddings interact with token-level representations.

\newpage

\appendix

\section*{Ethical Statement}

There are no ethical issues.




\bibliography{aaai25}

\end{document}